\documentclass{article}
\PassOptionsToPackage{table,xcdraw}{xcolor}

\usepackage[preprint]{neurips_2025}
\usepackage{neurips_2025}
\usepackage[utf8]{inputenc} 
\usepackage[T1]{fontenc}    
\usepackage{hyperref}       
\usepackage{url}            
\usepackage{booktabs}       
\usepackage{amsfonts}       
\usepackage{nicefrac}       
\usepackage{microtype}      
\usepackage{xcolor}         
\usepackage{graphicx}
\usepackage[]{natbib}
\usepackage{amsthm}  
\usepackage{amsmath}
\usepackage{changepage}
\usepackage{makecell}
\usepackage{amssymb}
\usepackage{multirow}
\usepackage{array}
\usepackage{xcolor}
\definecolor{myblue}{RGB}{235,238,255}
\definecolor{myred}{RGB}{255,240,240}

\definecolor{mygreen}{RGB}{239,255,253}
\definecolor{myyellow}{RGB}{235,255,235}
\definecolor{myorange}{RGB}{254,245,236}

\definecolor{app1}{RGB}{253,255,243}
\definecolor{app2}{RGB}{252,246,251}
\definecolor{app3}{RGB}{254,244,244}
\definecolor{app4}{RGB}{243,247,255}

\newtheorem{Observation}{Observation}

\title{Towards Robust Spiking Neural Networks:\\Mitigating Heterogeneous Training Vulnerability via\\ Dominant Eigencomponent Projection}

\author{%
  Desong Zhang\qquad Jia Hu\qquad Geyong Min \\
  Department of Computer Science, University of Exeter\\
  Exeter EX4 4RN, United Kingdom \\
  \texttt{\{dz288, J.Hu, G.Min\}@exeter.ac.uk} \\
}

\begin{document}

\maketitle

\begin{abstract}
Spiking Neural Networks (SNNs) process information via discrete spikes, enabling them to operate at remarkably low energy levels. However, our experimental observations reveal a striking vulnerability when SNNs are trained using the mainstream method—direct encoding combined with backpropagation through time (BPTT): even a single backward pass on data drawn from a slightly different distribution can lead to catastrophic network collapse. Our theoretical analysis attributes this vulnerability to the repeated inputs inherent in direct encoding and the gradient accumulation characteristic of BPTT, which together produce an exceptional large Hessian spectral radius. To address this challenge, we develop a hyperparameter-free method called \textbf{D}ominant \textbf{E}igencomponent \textbf{P}rojection (DEP). By orthogonally projecting gradients to precisely remove their dominant components, DEP effectively reduces the Hessian spectral radius, thereby preventing SNNs from settling into sharp minima. Extensive experiments demonstrate that DEP not only mitigates the vulnerability of SNNs to heterogeneous data poisoning, but also significantly enhances overall robustness compared to key baselines, providing strong support for safer and more reliable SNN deployment.
  
\end{abstract}







\section{Introduction}

As an emerging brain-inspired computational paradigm, Spiking Neural Networks (SNNs) leverage event-driven, discrete spike streams for feature representation \citep{maass1997networks}. By eliminating the need for pervasive and computationally intensive matrix multiplications of traditional Artificial Neural Networks (ANNs), SNNs achieve remarkable computational efficiency and significantly lower energy consumption \citep{pei2019towards, meng2023towards}. Owing to these inherent advantages, SNNs have been applied across a diverse array of application domains, such as autonomous driving \citep{zhu2024autonomous, shalumov2021lidar, viale2021carsnn}, edge computing \citep{liu2024energy, zhang2024spiking}, medical image process \citep{liu2025ssefusion, pan2024eg}, and robot control \cite{jiang2025fully, kumar2025dsqn}.

\begin{figure}
  \centering
  \includegraphics[width=1\linewidth]{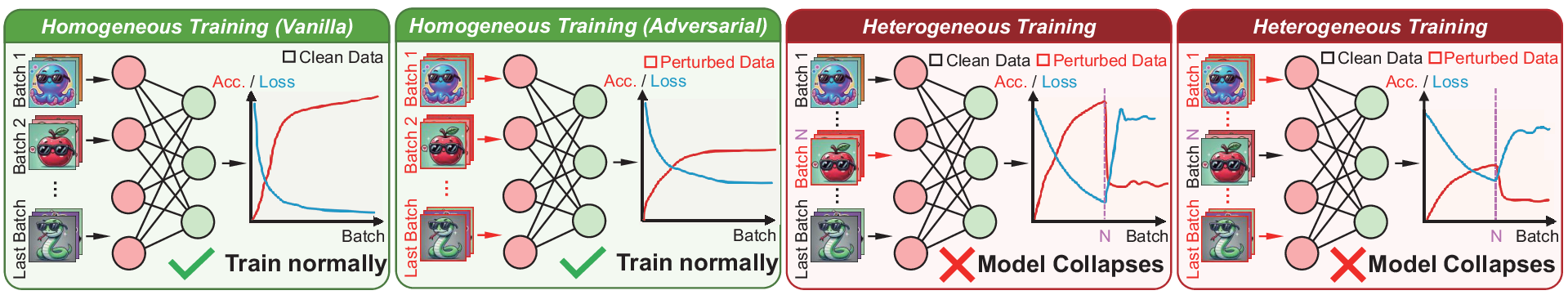}
  \caption{The vulnerability of SNNs in heterogeneous training.}
  \label{fig1}
\end{figure}

In the practical deployment of SNNs, safety and reliability are of paramount importance, particularly in terms of robustness against perturbations. Even subtle perturbations in the input data that are imperceptible to human senses  can trigger severely adverse and unpredictable network responses \citep{ding2024robust}. To enhance the robustness of SNNs, existing studies predominantly adopt a homogeneous training paradigm, where models are trained on data drawn from a single, uniform distribution—for instance, vanilla training using only clean samples \citep{ding2024enhancing, ding2024robust, geng2023hosnn, ding2022snn}, or adversarial training where all inputs are perturbed with equal intensity \citep{ding2024enhancing, geng2023hosnn, liu2024enhancing}. However, such training settings are idealized and do not reflect the variability and complexity of real-world data. In practical scenarios, models are often required to learn from inherently unpredictable and heterogeneous data distributions, as adversaries may employ a wide range of poisoning strategies to deliberately disrupt distributional homogeneity. We refer to this more realistic paradigm as heterogeneous training  (hetero-training). Notably, from the perspective of the attacker, when the number of manipulable samples is limited, concentrating these perturbed samples as a batch—rather than dispersing them sporadically throughout the dataset—often leads to a more pronounced degradation of model performance \citep{zou2022defending}. When exposed to batch-level heterogeneity in the training data, we observe a critical vulnerability in SNNs as follows:

\begin{Observation}
In SNN training phase, even a single backward pass with a slightly differently-distributed batch can trigger complete model collapse. As depicted in Fig. \ref{fig1}, SNNs trained on homogeneous datasets—whether comprised solely of clean samples or perturbed ones—exhibit a stable training trajectory. However, introducing just one batch of perturbed data into a clean dataset, or vice versa, leads to abrupt and catastrophic model collapse. \textbf{We refer to this phenomenon as the heterogeneous training vulnerability of SNNs.} (Sec. \ref{method1} presents a comprehensive analysis of the experimental results regarding the Observation 1.)
\end{Observation}

This observation reveals a fundamental security risk in SNNs when dealing with training data that is inherently unpredictable and cannot be predefined—a scenario often encountered in real-world adversarial contexts. This prompts the following questions:

\framebox[1\textwidth][l]{%
  \parbox{0.98\textwidth}{%
    \textbf{1.} Why do SNNs experience model collapse in hetero-training? \\
    \textbf{2.} Without relying on input data manipulation, can we design an approach for SNNs that effectively mitigates the model collapse induced by hetero-training and enhance robustness?
  }
}

Motivated by these questions, we propose a novel training method that enhances the robustness of SNNs under both homogeneous and heterogeneous training conditions. Specifically, 

\begin{itemize}
    \item  We theoretically analyze the root cause of SNN model collapse, demonstrating that mainstream SNN training methods—characterized by repeated inputs and gradient accumulation—induce extremely large spectral radius in the Hessian matrix.
    \item Building on these theoretical insights, we develop a  hyperparameter-free Dominant Eigencomponent Projection (DEP) method. By orthogonally projecting the gradients to precisely eliminate their dominant components, DEP effectively reduces the Hessian spectral radius, thereby preventing the network from falling into sharp minima.
    \item Extensive experimental results demonstrate that DEP mitigates SNN vulnerabilities and significantly enhances robustness under both homogeneous and heterogeneous training conditions, outperforming key baselines and thereby ensuring greater safety and reliability in deployment.
\end{itemize}

\section{Background and Related Work}
\subsection{Spiking Neuron Dynamic}
In SNNs, neurons emulate the spiking behavior of biological neurons to facilitate information transmission. One of the most prevalent nonlinear spiking neuron models in SNNs is the Leaky Integrate-and-Fire (LIF) neuron \cite{xu2022hierarchical, fang2021deep, ding2022snn, kundu2021hire}. The dynamics of a LIF neuron are described by Eq. (\ref{lif}), where \(x_t\), \(V_t\), and \(S_t\) represent the input, membrane potential, and spike output at time \(t\), respectively. Here, \(\tau\) denotes the membrane time constant, \(V_{\text{th}}\) is the potential threshold, and \(\Theta\) corresponds to the Heaviside function.
\begin{equation}
\label{lif}
    \tau\frac{dV_t}{dt} = -V_t+x_t, \quad S_t=\Theta(V_t-V_{\text{th}})
\end{equation}

\subsection{Adversarial Attack}
%


Given an input \(x\) with label \(y\), adversarial examples are generated by finding a perturbation \(\delta\) that maximizes the loss \(\mathcal{L}(h(x+\delta), y)\). This optimization problem is formally expressed as:

\begin{equation}
\underset{\|\delta\|_p \le \epsilon}{\arg\max}\; \mathcal{L}(f(x+\delta), y).
\end{equation}
In this paper, we employ two widely adopted gradient-based adversarial attack methods—Fast Gradient Sign Method (FGSM) \cite{goodfellow2014explaining} and Projected Gradient Descent method (PGD) \cite{madry2017towards}. Detailed descriptions and parameter configurations for these techniques are provided in  Appendix \ref{App_attack}.

\subsection{Related Work of SNN Defensive Methods}

As the most widely adopted training approach, direct training based on Backpropagation Through Time (BPTT) \citep{wu2018spatio} is extensively employed for SNNs. Early research in this domain typically utilized rate encoding to process input data, a technique that has been shown to endow models with inherent robustness against adversarial attacks \cite{kim2022rate, sharmin2020inherent, sharmin2019comprehensive}. However, rate encoding generally requires multiple time steps to fully capture the information, which spurred the development of direct encoding method that can operate with as few as a single time step \citep{rueckauer2017conversion}. Consequently, the combination of direct encoding with BPTT has emerged as the predominant training paradigm for SNNs \cite{wu2018spatio, deng2022temporal, meng2023towards, xiao2022online, wang2023ssf}. While direct encoding enhances both training and inference efficiency, it unfortunately forfeits the intrinsic robustness that rate encoding naturally provides. Based on direct encoding and BPTT, several approaches have been proposed to enhance the robustness of SNNs. For instance, \cite{kundu2021hire} introduces meticulously designed noise into the input, while \cite{ding2022snn} develops an adversarial training scheme regularized via Lipschitz analysis. In addition, \cite{liu2024enhancing} implements gradient sparsity regularization, \cite{xu2024feel} proposes an evolutionary leak factor for neuronal dynamics, and \cite{ding2024robust} introduces a modified training framework aimed at reducing the mean square of membrane potential perturbations. However, these studies have not explored why the combination of direct encoding and BPTT renders SNNs so susceptible to even minor perturbations from a mechanistic standpoint. In this paper, we leverage the observed phenomenon of model collapse to conduct an in-depth analysis of SNN vulnerability and introduce our DEP method to enhance network robustness.





\section{Analysis and Method}

In this section, we first present experimental results that demonstrate the model collapse phenomenon described in \textit{Observation 1} and analyze why SNNs exhibit a disconcerting vulnerability in hetero-training (Sec. \ref{method1}). Building on this analysis, we introduce the Dominant Eigencomponent Projection method and subsequently provide a theoretical analysis explaining how our approach mitigates the vulnerabilities associated with hetero-training while simultaneously enhancing network robustness (Sec. \ref{method2}).

\subsection{Preliminary Experiment and Analysis: Why SNN Model Collapses in Hetero-training?}
\label{method1}

\begin{figure}
  \centering
  \includegraphics[width=1\linewidth]{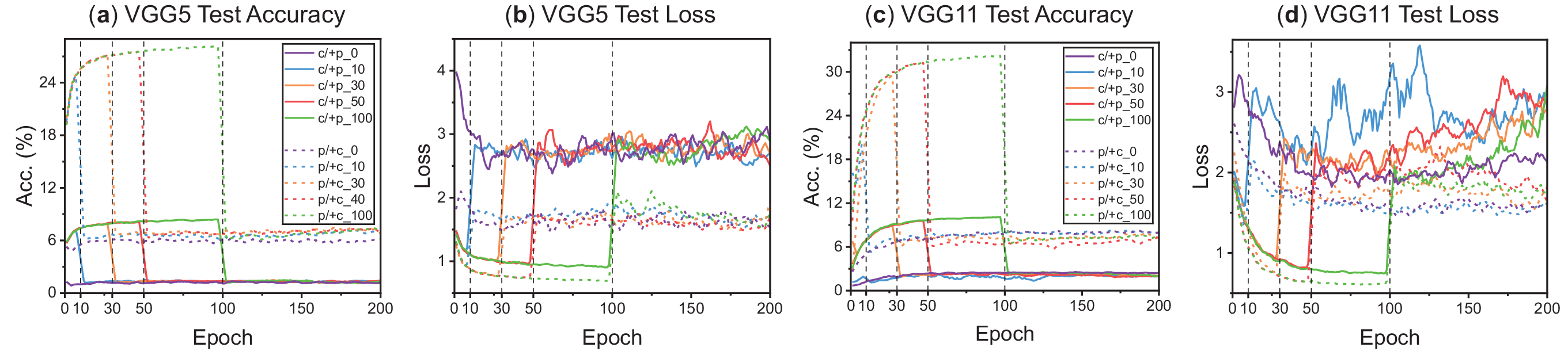}
  \caption{SNN model collapse curves caused by hetero-training. In the "c/+p\_10" setting, the model undergoes homogeneous training on a clean CIFAR-10 dataset for the first 9 epochs, after which hetero-training is initiated starting from epoch 10. During the heterogeneous phase, one batch per epoch is randomly selected and poisoned using perturbed data (FGSM with $\epsilon$ = 2). In the "p/+c" setting, homogeneous training is performed using perturbed data (FGSM with $\epsilon$ = 2), and during the heterogeneous phase, one random batch per epoch is poisoned with clean data.}
  \label{fig2}
\end{figure}

Fig. \ref{fig2} presents the complete training curves illustrating the phenomenon described in \textit{Observation 1}. In these experiments, we monitored both accuracy and loss on the CIFAR-10 test set using two VGG models of varying depth: VGG5 (Figures (\textbf{a}) and (\textbf{b})) and VGG11 (Figures (\textbf{c}) and (\textbf{d})). It is evident that for the shallower VGG5 network, whether hetero-training is applied throughout (c/+p\_0 and p/+c\_0) or abruptly introduced at any epoch (c/+p\_10-100, p/+c\_10-100), the network invariably collapses to a similarly unacceptable performance level, with loss trajectories that nearly coincide. Similarly, for the deeper VGG11 model, while the accuracy and loss curves corresponding to hetero-training at different epochs do not align as closely as in VGG5, they nonetheless exhibit a fundamentally analogous trend—namely, that hetero-training leads to a severe collapse in model performance.

This phenomenon suggests that, within a single epoch, network parameters may intermittently reside in extremely sharp local minima—characterized by an abnormally large spectral radius of the Hessian of the loss function \citep{cheng2022adversarial}. To explain this occurrence, we analyze it in the context of SNN properties.

\textbf{Impact of BPTT on Hessian Eigenvalues.} Consider an SNN trained via BPTT on input sequences $\{x_t\}_{t=1}^T$ over $T$ discrete time steps. We denote the parameter set by $\theta$, and let $\mathcal{L}(\theta)$ be the cumulative loss function with respect to parameters $\theta$. Concretely,

\begin{equation}
    \mathcal{L}(\theta) = \sum_{t=1}^{T} \ell\bigl(y_t, \hat{y}_t(\theta)\bigr),
\end{equation}

The Hessian matrix $H(\theta)$ is defined as the second derivative of $\mathcal{L}(\theta)$:

\begin{equation}
    H(\theta) = \nabla_{\theta}^2 \mathcal{L}(\theta) = \sum_{t=1}^{T} \frac{\partial^2 \ell\bigl(y_t,\hat{y}_t\bigr)}{\partial \hat{y}_t^2}
\left(\frac{\partial \hat{y}_t}{\partial \theta}\right)
\left(\frac{\partial \hat{y}_t}{\partial \theta}\right)^{\top}
\;+\;
\sum_{t=1}^{T} \frac{\partial \ell\bigl(y_t,\hat{y}_t\bigr)}{\partial \hat{y}_t}
\frac{\partial^2 \hat{y}_t}{\partial \theta^2}.
\end{equation}

BPTT computes gradients by unfolding the network across time. Consequently, the gradient with respect to parameters at each time step accumulates multiplicatively through the chain rule:

\begin{equation}
\frac{\partial \hat{y}_t}{\partial \theta}
= \sum_{k=1}^{t} \frac{\partial \hat{y}_t}{\partial x_{k}} \,\frac{\partial x_{k}}{\partial \theta}.
\end{equation}

Here, the gradients at time $t$ explicitly depend on gradients at previous time steps $k<t$, causing multiplicative accumulation across time steps. As a result, gradients can exhibit exponential amplification:


\begin{equation}
\frac{\partial \hat{y}_t}{\partial \theta}
\approx \prod_{k=1}^{t} \frac{\partial x_{k}}{\partial x_{k-1}} \,\frac{\partial x_{1}}{\partial \theta}.
\end{equation}

This exponential accumulation of gradients contributes significantly to the magnitude of Hessian entries, especially through the rank-1 structure of $\left(\frac{\partial \hat{y}_t}{\partial \theta}\right)\left(\frac{\partial \hat{y}_t}{\partial \theta}\right)^{\top}$. Hence, we have the following proportional relationship for Hessian’s spectral radius due to BPTT:

\begin{equation}
\rho\bigl(H(\theta)\bigr)
\,\propto\,
\sum_{t=1}^{T}\left\|\frac{\partial \hat{y}_t}{\partial \theta}\right\|^{2}
\,\approx\,
T \,\cdot\, \exp\bigl(2T \,\bar{\lambda}_J\bigr).
\end{equation}

where $\bar{\lambda}_J$ denotes the average log-magnitude eigenvalue of the Jacobian $\frac{\partial x_{t}}{\partial x_{t-1}}$. This indicates that the multiplicative accumulation of gradients induced by BPTT amplifies the spectral radius of the Hessian by a factor on the order of \(T\exp(2T)\) as the number of time steps \(T\) increases, thus creating a large spectral gap between $\rho\bigl(H\bigr)$ and the remaining eigenvalues, thereby reinforcing its dominant influence on the curvature of the loss landscape.

\textbf{Additional Amplification from Direct Encoding.} The direct encoding method further exacerbates this phenomenon. Under direct encoding, the input signal at each time step is constant \citep{rueckauer2017conversion}, which means $\forall t$, we have $x_t = x$. Consequently, the input Jacobian $J_x(\theta)=\frac{\partial x}{\partial \theta}$ is identical across all time steps, causing each time step’s gradient to reinforce identical parameter directions repeatedly. Mathematically, we have:

\begin{equation}
\frac{\partial \hat{y}_t}{\partial \theta}
= \sum_{k=1}^{t} \frac{\partial \hat{y}_t}{\partial x_{k}} \, J_{x}(\theta).
\end{equation}

Given the identical input across all time steps, the Hessian contribution simplifies to:

\begin{equation}
H(\theta)
\approx
T \cdot J_{x}(\theta)\,
\Bigl(\frac{\partial^2 \ell}{\partial \hat{y}^2}\Bigr)_{\text{avg}}\,
J_{x}(\theta)^{\top}.
\end{equation}

Thus, the spectral radius of the Hessian grows proportionally to the number of time steps $T$:

\begin{equation}
\rho\bigl(H(\theta)\bigr)
\approx
T \cdot
\lambda_{\max}\!\Bigl(
J_{x}(\theta)\,
\bigl(\frac{\partial^2 \ell}{\partial \hat{y}^2}\bigr)\,
J_{x}(\theta)^{\top}
\Bigr).
\end{equation}

This demonstrates that direct encoding intensifies Hessian eigenvalue growth by repeatedly emphasizing identical gradient directions.

Combining BPTT and direct encoding yields a compounded effect: (i) BPTT induces exponential gradient accumulation over time, and (ii) direct encoding imposes repeated identical input, strengthening certain parameter directions exponentially with respect to time steps. Together, these two factors result in inevitably severe amplification of Hessian eigenvalues in one principal direction as Eq. (\ref{final}), yielding an excessively large second-order loss component. This Hessian sharpness significantly destabilizes training by causing even small perturbations to result in disproportionately high loss increments, which explains why SNNs being highly sensitive to small drifted distributions, such as introducing heterogeneous batch, and thus easily collapsing to sharp minima as \textit{Observation 1} shows.

\begin{equation}
\label{final}
\rho\bigl(H(\theta)\bigr)
\,\sim\,
T \,\cdot\, \exp\!\bigl(2T \,\bar{\lambda}_J\bigr)
\,\cdot\,
\lambda_{\max}\!\Bigl(
J_{x}(\theta)\,\bigl(\frac{\partial^2 \ell}{\partial \hat{y}^2}\bigr)J_{x}(\theta)^{\top}
\Bigr).
\end{equation}

\subsection{Reducing Spectral Radius via Dominant Eigencomponent Projection}
\label{method2}

To alleviate this pathological eigenvalue amplification—that is, to reduce the spectral radius of the Hessian—we propose a deterministic and hyperparameter-free gradient update technique named Dominant Eigencomponent Projection (DEP), whose essence is to remove the contribution of the dominant eigencomponent from the overall gradient via an projection operation. 

Formally, we denote the gradient of loss function $\mathcal{L}(\theta)$ as a general $k$-dimensional tensor, that is $\nabla_{\theta}\mathcal{L}(\theta) \in \mathbb{R}^{d_1 \times d_2 \times \cdots \times d_k}$. To systematically analyze and manipulate the principal components of this high-dimensional gradient, we first define a deterministic matrixization operator $\mathcal{M}$ as follows:

\begin{equation}
    \mathcal{M}: \mathbb{R}^{d_1 \times d_2 \times \cdots \times d_k} \rightarrow \mathbb{R}^{m \times n}, \quad \text{with} \quad m=d_1,\quad n=\prod_{j=2}^{k}d_j,
\end{equation}

such that explicitly, we have $\mathcal{M} \bigl(\nabla_{\theta}\mathcal{L}(\theta)  \bigr)\in \mathbb{R}^{m \times n}$. Next, we perform a deterministic singular value decomposition on $\mathcal{M} \bigl(\nabla_{\theta}\mathcal{L}(\theta)\bigr)$:

\begin{equation}
\mathcal{M} \bigl(\nabla_{\theta}\mathcal{L}(\theta)\bigr)
= U \Sigma V^{\top}
= \sum_{i=1}^{r} \sigma_{i}\,u_{i}\,v_{i}^{\top},
\quad
r = \min(m,n), \quad \sigma_{1} \ge \sigma_{2} \ge \cdots \ge \sigma_{r} \ge 0.
\end{equation}

We define the dominant eigencomponent (in the sense of the gradient’s dominant singular contribution) as the outer product of leading singular vectors $u_1,v_1$, that is, $u_1v_1^{\top}\in\mathbb{R}^{m\times n}$. To remove this dominant eigencomponent from the gradient, we introduce the projection operator $\mathcal{P}_{u_1v_1^{\top}}(A)$ defined for any matrix $A\in\mathbb{R}^{m\times n}$ as Eq. (\ref{projection}) with the Frobenius inner product: $\langle A,B\rangle_{F} = \sum_{i=1}^m\sum_{j=1}^nA_{ij}B_{ij}$, and Frobenius norm $\|A\|_{F}=
\sqrt{\langle A,A\rangle_{F}}$.

\begin{equation}
\label{projection}
\mathcal{P}_{u_1v_1^{\top}}(A)
\;=\;
\frac{\langle A,\;u_1v_1^{\top}\rangle_{F}}{\|u_1v_1^{\top}\|_{F}^{2}}
\,u_1v_1^{\top}.
\end{equation}

Thus, DEP explicitly performs an orthogonal projection of $\mathcal{M}\bigl( \nabla_{\theta}\mathcal{L}(\theta) \bigr)$ onto the orthogonal complement of the dominant eigencomponent $u_1v_1^{\top}$. In other words, we arrive at the update rule as Eq. (\ref{method_function}), which deterministically projects out the dominant eigencomponent from the gradient.

\begin{equation}
\label{method_function}
\tilde{\nabla_{\theta}\mathcal{L}(\theta)}
= \mathcal{M}^{-1}\biggl(
\mathcal{M}\bigl( \nabla_{\theta}\mathcal{L}(\theta) \bigr)
-\;
\mathcal{P}_{u_1v_1^{\top}}\bigl(\mathcal{M}\bigl( \nabla_{\theta}\mathcal{L}(\theta) \bigr)\bigr)
\biggr).
\end{equation}

Consider the Hessian matrix at the parameter point $\theta$, denoted as $H(\theta)\in\mathbb{R}^{mn\times mn}$, and its eigendecomposition:
\begin{equation}
H(\theta)
= \sum_{i=1}^{mn} \lambda_{i}\,q_{i}\,q_{i}^{\top},
\quad
\lambda_{1} > \lambda_{2} \,\ge\, \cdots \,\ge\, \lambda_{mn}.
\end{equation}

In standard gradient descent, the local sharpness induced by gradient update $\text{vec}\bigl(\mathcal{M}\bigl( \nabla_{\theta}\mathcal{L}(\theta) \bigr) \bigr)$ approximately equals:

\begin{equation}
\kappa_{\mathrm{std}}
= \frac{\text{vec}\bigl(\mathcal{M}\bigl( \nabla_{\theta}\mathcal{L}(\theta) \bigr) \bigr)^{\top}H(\theta)\,\text{vec}\bigl(\mathcal{M}\bigl( \nabla_{\theta}\mathcal{L}(\theta) \bigr) \bigr)}{\|\text{vec}\bigl(\mathcal{M}\bigl( \nabla_{\theta}\mathcal{L}(\theta) \bigr) \bigr)\|_{2}^{2}}
\,\approx\,
\lambda_{1}.
\end{equation}

This scenario is particularly pronounced when the gradient is closely aligned with the principal Hessian eigenvector $u_1v_1^{\top}$. In this case, standard gradient updates suffer from excessive sharpness characterized by $\kappa_{\mathrm{std}}\approx \lambda_{1}$. By contrast, after the DEP update, the resulting gradient direction $\tilde{\text{vec}\bigl(\mathcal{M}\bigl( \nabla_{\theta}\mathcal{L}(\theta) \bigr) \bigr)}$ is explicitly projected orthogonal to $u_1v_1^{\top}$, rigorously ensuring a curvature bound given by Eq. (\ref{dep}).

\begin{equation}
\label{dep}
\kappa_{\mathrm{DEP}}
= \frac{\displaystyle\sum_{i=2}^{mn} \lambda_{i}\,\bigl((u_{i}v_{i}^{\top})^{\top}\tilde{\text{vec}\bigl(\mathcal{M}\bigl( \nabla_{\theta}\mathcal{L}(\theta) \bigr) \bigr)}\bigr)^{2}}
{\displaystyle\sum_{i=2}^{mn} \bigl((u_{i}v_{i}^{\top})^{\top}\tilde{\text{vec}\bigl(\mathcal{M}\bigl( \nabla_{\theta}\mathcal{L}(\theta) \bigr) \bigr)}\bigr)^{2}}
\;\le\;\lambda_{2}
< \lambda_{1}
\;\approx\;\kappa_{\mathrm{std}}.
\end{equation}

Thus, DEP strictly reduces the spectral radius of the Hessian, effectively addressing the pathological exponential eigenvalue amplification described previously. Besides, a reduction in Hessian spectral radius can indirectly lower the upper bound of the network's Lipschitz constant \citep{nesterov2013introductory, yao2020pyhessian, ghorbani2019investigation}, thereby contributing to improved robustness and enhanced generalization capabilities \citep{ding2022snn}.

\section{Experiment}

\subsection{Experimental Setup}
Our experiments are structured into four parts. First, we assess the robustness of DEP in homogeneous training (Sec. \ref{exp_sota}), which are divided into two settings: vanilla training using clean data and adversarial training (AT) \cite{kundu2021hire} using perturbed data generated by white-box FGSM with an \(\epsilon\) of \(2/255\). Second, we evaluate the effect of our approach on the Hessian eigenvalue during inference (Sec. \ref{exp_hessian}). Third, we investigate whether DEP can prevent catastrophic network collapse in hetero-training (Sec. \ref{exp_hetero}). Finally, we inspect our method for any instances of gradient obfuscation (Sec. \ref{exp_grad}). 

To ensure comprehensive evaluation, we conduct experiments on visual datasets of varying scales, including CIFAR-10 \cite{krizhevsky2009learning}, CIFAR-100 \cite{krizhevsky2009learning}, TinyImageNet \cite{deng2009imagenet}, and ImageNet \cite{deng2009imagenet}. Implementation specifics are provided in Appendix \ref{App_exp}.

\subsection{Comparison with State-of-the-art (SOTA) in Homogeneous Training}
\label{exp_sota}

\begin{figure}
  \centering
  \includegraphics[width=1\linewidth]{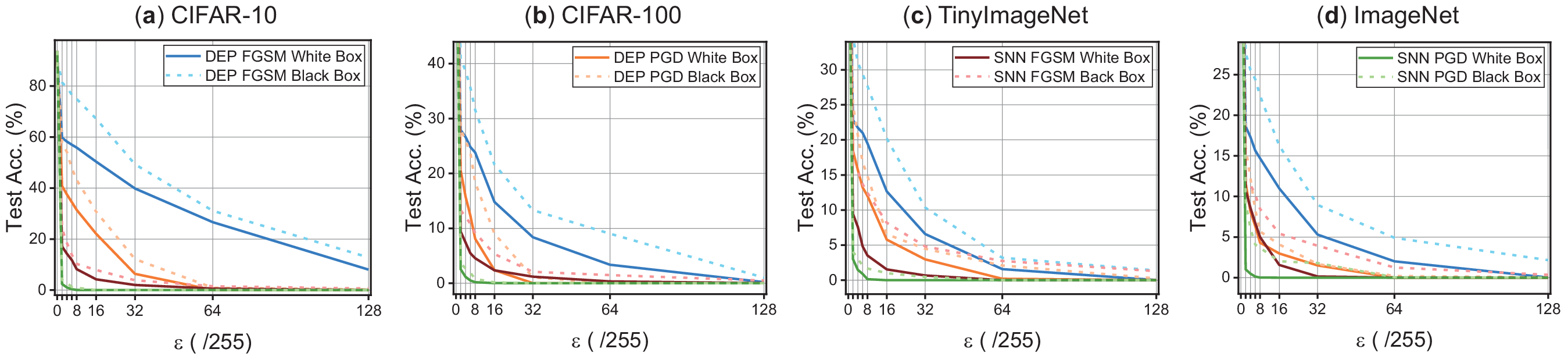}
  \caption{Performance comparison in different white box and black box attack.}
  \label{fig3}
\end{figure}

\textbf{White box attack.} 
Table \ref{tab1} summarizes the accuracies of DEP under various homogeneous training settings, compared against SOTA defenses (StoG \cite{ding2024enhancing}, DLIF \cite{ding2024robust}, HoSNN \cite{geng2023hosnn}, and FEEL \cite{xu2024feel}). Overall, DEP consistently delivers superior defense against gradient-based white‑box attacks across all datasets and training modes. Notably, DEP achieves more than 10\% accuracy improvements under several attack scenarios—including CIFAR‑100 with FGSM in vanilla training; CIFAR‑10 and ImageNet with FGSM and TinyImageNet with PGD in AT. In AT on TinyImageNet, DEP gains 30.87\% accuracy, representing an exciting 22.68\% margin above the baseline SNN’s 8.19\%.  On the other hand, bolstering robustness under vanilla training remains challenging, especially against PGD attacks. Despite this, DEP secures 8.09\% accuracy on CIFAR‑100 under PGD in the vanilla setting—a substantial breakthrough compared to the previous best of 2.04\% achieved by FEEL \cite{xu2024feel}.

\textbf{Black box attack.} 
Black box adversarial examples are generated using the substitute‑model approach in this experiment. We evaluated DEP’s resilience across a range of perturbation magnitudes and compared its performance to that of a vanilla SNN under the same attack strengths. Fig. \ref{fig3} visualizes these results, and the complete experimental data are provided in Appendix \ref{App_exp_res}. Across all four dashed‑line baselines, DEP consistently achieves substantially higher accuracy than the vanilla model, demonstrating its effectiveness against diverse black box threat scenarios.

\newcommand{\myrownum}{2}
\begin{table}[htbp]
\centering
\renewcommand{\arraystretch}{1.2} 
\caption{White box performance comparison (\%). The highest accuracy in each column is highlighted in bold. The “Improvement” quantifies the gain of DEP over the other best-performing baseline.}
\label{tab1}

\resizebox{1\textwidth}{!}{
\begin{tabular}{>{\centering\arraybackslash}m{2.0cm}|>{\centering\arraybackslash}m{0.05cm}|>{\centering\arraybackslash}m{1.1cm}>{\centering\arraybackslash}m{1.1cm}>{\centering\arraybackslash}m{1.1cm}|>{\centering\arraybackslash}m{0.05cm}|>{\centering\arraybackslash}m{1.1cm}>{\centering\arraybackslash}m{1.1cm}>{\centering\arraybackslash}m{1.1cm}|>{\centering\arraybackslash}m{0.05cm}|>{\centering\arraybackslash}m{1.1cm}>{\centering\arraybackslash}m{1.1cm}>{\centering\arraybackslash}m{1.1cm}|>{\centering\arraybackslash}m{0.05cm}|>{\centering\arraybackslash}m{1.1cm}>{\centering\arraybackslash}m{1.1cm}>{\centering\arraybackslash}m{1.1cm}}

\noalign{\hrule height 1.5pt}

  \multirow{2}{*}{\textbf{Methods}} & \multirow{\myrownum}{*}{} &
  \multicolumn{3}{c|}{\cellcolor{gray!20}\textbf{CIFAR-10}} & \multirow{\myrownum}{*}{} &
  \multicolumn{3}{c|}{\cellcolor{gray!20}\textbf{CIFAR-100}} & \multirow{\myrownum}{*}{} &
  \multicolumn{3}{c|}{\cellcolor{gray!20}\textbf{TinyImageNet}} & \multirow{\myrownum}{*}{} & 
  \multicolumn{3}{c}{\cellcolor{gray!20}\textbf{ImageNet}} \\

\cline{3-5}\cline{7-9}\cline{11-13} \cline{15-17}

  & & \cellcolor{gray!20}Clean & \cellcolor{gray!20}FGSM & \cellcolor{gray!20}PGD &
  & \cellcolor{gray!20}Clean & \cellcolor{gray!20}FGSM & \cellcolor{gray!20}PGD &
  & \cellcolor{gray!20}Clean & \cellcolor{gray!20}FGSM & \cellcolor{gray!20}PGD &
  & \cellcolor{gray!20}Clean & \cellcolor{gray!20}FGSM & \cellcolor{gray!20}PGD  \\

\hline
\multicolumn{17}{c}{\cellcolor{myblue}\textbf{Homogeneous Training: Vanilla Training}}\\
\hline

  SNN &  & \textbf{93.75}  & 8.19 & 0.03  &  &  72.39  & 4.55   &  0.19  &  & \textbf{56.82}  & 3.51  & 0.14  &  & \textbf{57.84}  & 4.99 & 0.01   \\
  \cline{1-1}\cline{3-5}\cline{7-9}\cline{11-13} \cline{15-17}
  StoG \cite{ding2024enhancing} &  & 91.64  & 16.22 & 0.28  &  & 70.44  & 8.27  & 0.49 &  &-  & - & - &  & - & - & - \\
  DLIF \cite{ding2024robust}&  & 92.22  & 13.24 & 0.09  &  & 70.79  & 6.95  & 0.08  &   &-  & - & - &  & - & - & -   \\
  HoSNN \cite{geng2023hosnn} &  & 92.43  & 54.76 & 15.32  &  & 71.98  & 13.48  & 0.19  &   &-  & - & - &  & - & - & -   \\
  FEEL \cite{xu2024feel} &  & 93.29  & 44.96 & 28.35  &  & \textbf{73.79}  & 9.60 & 2.04   &  & 43.83  & 9.59  &  4.53 &  & - &- &-  \\
  \cline{1-1}\cline{3-5}\cline{7-9}\cline{11-13} \cline{15-17}
  DEP (Ours) &  & 90.21  & \textbf{55.86} & \textbf{31.44}  &  & 70.26 & \textbf{23.81}  & \textbf{8.09}  &  & 54.54  & \textbf{19.50}  & \textbf{12.02}  &  & 53.47 & \textbf{14.69} & \textbf{4.30}  \\
  Improvement &  & \textcolor{blue}{-3.54}  & \textcolor{red}{+1.10} & \textcolor{red}{+3.09}  &  & \textcolor{blue}{-3.53}  & \textcolor{red}{+10.33}  & \textcolor{red}{+6.05}  &  & \textcolor{blue}{-2.28}  & \textcolor{red}{+9.91}   & \textcolor{red}{+7.49}  &  & \textcolor{blue}{-4.37}  & \textcolor{red}{+9.70} & \textcolor{red}{+4.29} \\

\hline
\multicolumn{17}{c}{\cellcolor{myred}\textbf{Homogeneous Training: Adversarial Training (AT)}}\\
\hline

  SNN &  & \textbf{91.16}  & 38.20 & 14.07 & & 69.69 & 16.31 & 8.49  &  & \textbf{49.91}  &8.19  &2.97  &  & \textbf{51.00}  &15.74  &6.39   \\
  \cline{1-1}\cline{3-5}\cline{7-9}\cline{11-13} \cline{15-17}
  StoG \cite{ding2024enhancing} &  & 90.13  & 45.74 & 27.74  &  & 66.37 & 24.45 & 14.42 &   &-  & - & - &  & - & - & -  \\
  DLIF \cite{ding2024robust}&  & 88.91  & 56.71 & 40.30  &  &66.33  & 36.83  & 24.25  &    &-  & - & - &  & - & - & - \\
  HoSNN \cite{geng2023hosnn} &  & 90.00  & 63.98 & 43.33 &  & 64.64  & 26.97 & 16.66  &   &-  & - & - &  & - & - & -   \\
  FEEL \cite{xu2024feel} &  & -  & - & -  &  & \textbf{69.79}  &  18.67 &  11.07 &    &-  & - & - &  & - & - & -  \\
  \cline{1-1}\cline{3-5}\cline{7-9}\cline{11-13} \cline{15-17}
  DEP (Ours) &  & 86.62  & \textbf{74.43} & \textbf{44.38}  &  & 64.21  & \textbf{43.91}  & \textbf{27.11}  &  & 46.30  &\textbf{30.87}  & \textbf{18.21}  & &49.78  & \textbf{26.83}  & \textbf{9.12}  \\
  Improvement &  & \textcolor{blue}{-4.54}  & \textcolor{red}{+10.45} & \textcolor{red}{+1.05}  &  & \textcolor{blue}{-5.58}  & \textcolor{red}{+7.08}  & \textcolor{red}{+2.86}  &  & \textcolor{blue}{-3.61}  & \textcolor{red}{+22.68} & \textcolor{red}{+15.24} &  & \textcolor{blue}{-1.22}  & \textcolor{red}{+10.09} & \textcolor{red}{+2.73} \\

\noalign{\hrule height 1.5pt}

\end{tabular}
}
\end{table}


\subsection{Hessian Eigenvalue Evaluation}
\label{exp_hessian}








In this experiment, we compare the Hessian eigenvalue of the DEP-trained model against those of a vanilla SNN during inference. Specifically, for each inference batch, we compute two metrics: (i). $\rho(H)$: The spectral radius of the Hessian; (ii). $\text{Pr}(H)$: The proportion of $\rho(H)$ within the top-five eigenvalues, serving as an indicator of the overall smoothness of the loss landscape.  The details for this experiment are provided in Appendix \ref{App_exp}. Table \ref{tab2} reports both metrics under three distinct adversarial attack scenarios. Across all cases, DEP consistently achieves a lower $\rho(H)$ and markedly reduces its proportional presence among the top-five eigenvalues. These findings corroborate our theoretical design, demonstrating that DEP indeed reduces the Hessian spectral radius, smooths the Hessian sharpness, and underpins its robustness enhancements.

\begin{table}[htbp]
\centering
\renewcommand{\arraystretch}{1.2} 
\caption{Hessian eigenvalue evaluation (Standard deviations). The smaller $\rho(H)$ and $\text{Pr}(H)$ are, the better. The situation where $\text{Pr}(H)$ is greater than 1 indicates that there are negative values among the top-five eigenvalues.}
\label{tab2}

\resizebox{1\textwidth}{!}{
\begin{tabular}{>{\centering\arraybackslash}m{0.8cm}|>{\centering\arraybackslash}m{0.05cm}|>{\centering\arraybackslash}m{2.7cm}>{\centering\arraybackslash}m{2.1cm}|>{\centering\arraybackslash}m{0.05cm}|>{\centering\arraybackslash}m{2.7cm}>{\centering\arraybackslash}m{2.1cm}|>{\centering\arraybackslash}m{0.05cm}|>{\centering\arraybackslash}m{2.7cm}>{\centering\arraybackslash}m{2.1cm}|>{\centering\arraybackslash}m{0.05cm}|>{\centering\arraybackslash}m{2.7cm}>{\centering\arraybackslash}m{2.1cm}}

\noalign{\hrule height 1.5pt}

  \multirow{2}{*}{\textbf{}} & \multirow{\myrownum}{*}{} &
  \multicolumn{2}{c|}{\cellcolor{gray!20}\textbf{CIFAR-10}} & \multirow{\myrownum}{*}{} &
  \multicolumn{2}{c|}{\cellcolor{gray!20}\textbf{CIFAR-100}} & \multirow{\myrownum}{*}{} &
  \multicolumn{2}{c|}{\cellcolor{gray!20}\textbf{TinyImageNet}} & \multirow{\myrownum}{*}{} & 
  \multicolumn{2}{c}{\cellcolor{gray!20}\textbf{ImageNet}} \\


  & & \cellcolor{gray!20} $\rho(H)$ & \cellcolor{gray!20} $\text{Pr}(H)$ & 
  & \cellcolor{gray!20}$\rho(H)$ & \cellcolor{gray!20}$\text{Pr}(H)$ & 
  & \cellcolor{gray!20}$\rho(H)$ & \cellcolor{gray!20}$\text{Pr}(H)$ & 
  & \cellcolor{gray!20}$\rho(H)$ & \cellcolor{gray!20}$\text{Pr}(H)$  \\

\hline
\multicolumn{13}{c}{\cellcolor{myyellow}\textbf{Clean Inference}}\\
\hline
SNN &  & 261.94 (78.28)  & 0.98 (5.92)    &  & 187.49 (47.35)   & 0.41 (0.06)      &  & 2077.09 (803.30)    & 0.60 (0.19)  &    & 174.88 (50.41)  & 0.43 (0.09)    \\
DEP &  & 209.90 (59.97)   & 0.35 (0.15) &  & 135.55 (36.26)   &  0.30 (0.04)    &  & 1802.80 (700.78)    & 0.51 (0.10)   &    & 110.56 (31.66)  & 0.33 (0.06)   \\

\hline
\multicolumn{13}{c}{\cellcolor{mygreen}\textbf{FGSM Inference}}\\
\hline

SNN &  & 269.90 (85.54)  & 1.15 (6.78)   &  & 190.82 (49.57)      & 0.46 (0.06)  &    & 1998.11 (741.19)  &  0.61 (0.05)   &  &162.57 (51.23)  & 0.44 (0.10)  \\
DEP &  & 218.27 (47.66)  & 0.38 (0.18)    &  &132.01 (30.17)   & 0.29 (0.04)      &  & 1767.57 (681.68)    & 0.50 (0.06)  &    &111.77 (27.40)  & 0.32 (0.06)    \\

\hline
\multicolumn{13}{c}{\cellcolor{myorange}\textbf{PGD Inference}}\\
\hline

SNN &  & 265.47 (74.16)  &  1.03 (6.34)   &  & 200.11 (45.24)   &  0.46 (0.05)    &  & 2072.77 (807.49)    & 0.62 (0.31)   &        & 174.16 (49.00) & 0.41 (0.06)   \\
DEP &  & 202.89 (63.24)  & 0.35 (0.10)    &  & 143.07 (37.96)  & 0.30 (0.04)     &  &1793.12 (733.90)     & 0.56 (0.34)   &        & 113.73 (27.97) & 0.32 (0.07)   \\

\noalign{\hrule height 1.5pt}
\end{tabular}
}
\end{table}

\subsection{Performance in Heterogeneous Training}
\label{exp_hetero}


Having confirmed DEP’s robustness in homogeneous training, we next evaluate its resilience in heterogeneous settings. We test poisoning intensities of $b = 1, 2, 5$ to examine whether DEP can avert catastrophic collapse under increasing poison strength. The results, including the performance degeneration, detailed in Table \ref{tab3}, omit the vanilla SNN baselines due to their near-total failure under heterogeneous poisoning. Remarkably, DEP maintains strong resilience across all settings, with no instance of full collapse. Even at the highest poisoning strength ($b = 5$), DEP sustains a 30\% accuracy on CIFAR‑10 under FGSM inference, demonstrating its capacity to withstand realistic, batch‑level data poisoning.

\begin{table}[htbp]
\centering
\renewcommand{\arraystretch}{1.2} 
\caption{DEP performance in hetero-training (\%). Following the protocol of Sec. \ref{method1}, we define two poisoning schemes: c/+p\_0 and p/+c\_0. In c/+p\_0, training with clean data, beginning at epoch 0, we inject $b$ FGSM ($\epsilon = 2/255$)–perturbed batches at the end of each epoch; p/+c\_0 is defined analogously: trained with perturbed data and injected with clean data. The data in parentheses represents the difference from the baseline ($b=0$).}
\label{tab3}

\resizebox{1\textwidth}{!}{
\begin{tabular}{>{\centering\arraybackslash}m{0.8cm}|>{\centering\arraybackslash}m{0.05cm}|>{\centering\arraybackslash}m{2.3cm}>{\centering\arraybackslash}m{2.3cm}>{\centering\arraybackslash}m{2.3cm}>{\centering\arraybackslash}m{2.3cm}|>{\centering\arraybackslash}m{0.05cm}|>{\centering\arraybackslash}m{2.3cm}>{\centering\arraybackslash}m{2.3cm}>{\centering\arraybackslash}m{2.3cm}>{\centering\arraybackslash}m{2.3cm}}

\noalign{\hrule height 1.5pt}

  \multirow{2}{*}{\textbf{$b$}} & \multirow{\myrownum}{*}{} &
  \multicolumn{4}{c|}{\cellcolor{gray!20}\textbf{Hetero-training: c/+p\_0}} & \multirow{\myrownum}{*}{} &
  \multicolumn{4}{c}{\cellcolor{gray!20}\textbf{Hetero-training: p/+c\_0}} \\


  & & \cellcolor{gray!20}CIFAR-10 &  \cellcolor{gray!20}CIFAR-100  
  & \cellcolor{gray!20}TinyImageNet & \cellcolor{gray!20}ImageNet & 
  & \cellcolor{gray!20}CIFAR-10 & \cellcolor{gray!20}CIFAR-100  
  & \cellcolor{gray!20}TinyImageNet &  \cellcolor{gray!20}ImageNet \\

\hline
\multicolumn{11}{c}{\cellcolor{myyellow}\textbf{Clean Inference}}\\
\hline

0 &  & 90.21            & 70.26            & 54.54            & 53.47   &   & 86.62            & 64.21            & 46.30            & 49.78            \\
1 &  & 40.58 (-49.63)   & 27.40 (-42.86)   & 21.45 (-33.09)   & 23.54 (-29.93)  &   & 38.61 (-48.01)   & 22.08 (-42.13)   & 19.30 (-27.00)   & 22.69 (-27.09)   \\
2 & & 31.08 (-59.13)   & 12.76 (-57.50)   &  9.64 (-44.90)   & 10.53 (-42.94)  &   & 29.95 (-56.67)   & 11.58 (-52.63)   &  8.95 (-37.35)   & 9.03 (-40.75)   \\
5 & & 12.07 (-78.14)  &  4.18 (-66.08)   &  3.98 (-50.56)   &  5.67 (-47.80)  &   & 12.02 (-74.60)   &  3.77 (-60.44)   &  2.72 (-43.58)   &  3.44 (-46.34)   \\

\hline
\multicolumn{11}{c}{\cellcolor{mygreen}\textbf{FGSM Inference}}\\
\hline
0 &  & 55.86            & 23.81            & 19.50            & 14.59  &   & 74.43            & 43.91            & 30.87            & 26.83            \\
1 &  & 33.70 (-22.16)   & 12.54 (-11.27)   & 14.67 (-4.83)    & 11.45 (-3.14)  &   & 40.09 (-34.34)   & 19.16 (-24.75)   & 12.04 (-18.83)   & 11.68 (-15.15)   \\
2 &  & 33.29 (-22.57)   & 13.70 (-10.11)   & 14.78 (-4.72)    & 11.12 (-3.47)  &   & 38.22 (-36.21)   & 17.31 (-26.60)   & 10.11 (-20.76)   &  8.64 (-18.19)   \\
5 &  & 34.85 (-21.01)   & 14.93 (-8.88)    & 15.56 (-3.94)    & 11.86 (-2.73)  &   & 31.58 (-42.85)   &  8.49 (-35.42)   &  5.80 (-25.07)   &  3.10 (-23.73)   \\

\hline
\multicolumn{11}{c}{\cellcolor{myorange}\textbf{PGD Inference}}\\
\hline
0 &  & 31.44            &  9.09            & 12.02            &  4.30  &   & 44.38            & 27.11            & 18.21            &  9.12            \\
1 &  & 22.87 (-8.57)    &  3.97 (-5.12)    &  8.89 (-3.13)    &  2.67 (-1.63)  &   & 20.77 (-23.61)   & 13.71 (-13.40)   & 12.44 (-5.77)    &  4.90 (-4.22)    \\
2 &  & 21.23 (-10.21)   &  3.34 (-5.75)    &  8.67 (-3.35)    &  2.21 (-2.09)  &   & 18.72 (-25.66)   & 11.22 (-15.89)   & 11.90 (-6.31)    &  3.38 (-5.74)    \\
5 &  & 21.69 (-9.75)    &  3.01 (-6.08)    &  9.02 (-3.00)    &  1.88 (-2.42)  &   & 14.31 (-30.07)   &  7.93 (-19.18)   &  8.78 (-9.43)    &  1.94 (-7.18)    \\

\noalign{\hrule height 1.5pt}
\end{tabular}
}
\end{table}

\subsection{Inspection of Gradient Obfuscation}
\label{exp_grad}

The seminal work \cite{athalye2018obfuscated} critically identified several characteristic behaviors—summarized in Table 4—that arise when a defense attains spurious robustness through gradient obfuscation. Accordingly, we evaluate DEP against each of these behaviors. Our experiments demonstrate that DEP passes all checks, as detailed below:

For items (1) and (2), Fig. \ref{fig3} presents DEP’s accuracy under FGSM and PGD attacks across a range of perturbation bounds, as well as a side‑by‑side comparison of white‑box versus black‑box performance. It is clear that DEP is consistently more vulnerable to iterative PGD than to single‑step FGSM, and that white‑box attacks inflict greater degradation than black‑box attacks.  Items (3) and (4) are also evident in Fig. \ref{fig3}: as the perturbation limit increases, DEP’s accuracy drops sharply, even reaching 0\% under several settings. Fig. 4 further corroborates this trend, showing that although DEP’s performance progressively worsens with more PGD iterations, it eventually converges to a steady minimum.  Finally, item (5) indicates that gradient‑based attacks fail to locate adversarial examples; however, our results in Fig. \ref{fig3} demonstrate the opposite—both FGSM and PGD continue to fool DEP despite the training.

\begin{figure*}[ht]
    \begin{minipage}[t]{0.695\textwidth}
    \centering
    \label{tab4}
    \renewcommand{\arraystretch}{1.2} 
    Table 4: Checklist for identifying gradient obfuscation \\
    
    \vspace{0.2cm}
    
    \resizebox{1\textwidth}{!}{
    \begin{tabular}{>{\arraybackslash}m{9.5cm}>{\centering\arraybackslash}m{1cm}}
    
    \noalign{\hrule height 1.5pt}

    \multicolumn{1}{c}{ \cellcolor{gray!20}Characteristics to identify gradient obfuscation} &  \multicolumn{1}{c}{ \cellcolor{gray!20}Pass?}\\
    
    \hline

    (1) Single-step attack performs better compared to iterative attacks & \checkmark \\
    (2) Black-box attacks perform better compared to white-box attacks & \checkmark \\
    (3) Increasing perturbation bound can’t increase attack strength & \checkmark \\
    (4) Unbounded attacks can’t reach 100\% success & \checkmark \\
    (5) Adversarial example can be found through random sampling & \checkmark \\

    \noalign{\hrule height 1.5pt}
    \end{tabular}
    }

    \end{minipage}%
    \hfill
    \begin{minipage}[t]{0.305\textwidth}
    \vspace{1pt}
    \includegraphics[width=1\columnwidth]{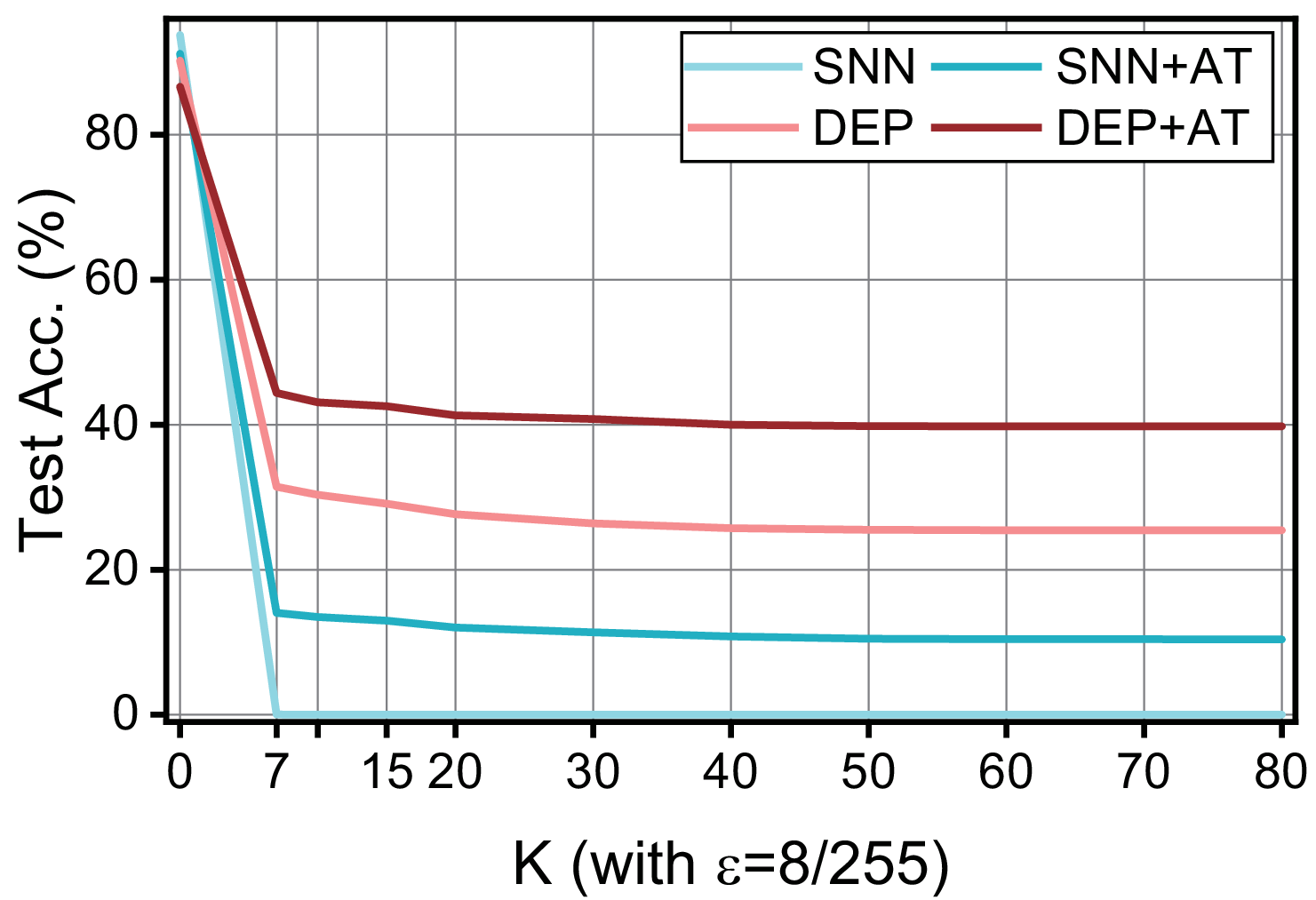} 
    Figure 4: Acc. in different $K$.
    
    \label{fig_LS}

    \end{minipage}
\end{figure*}

\setcounter{table}{3} 
\refstepcounter{table} 

\setcounter{figure}{3} 
\refstepcounter{figure} 


%





\section{Conclusion and Discussion}
\label{conclusion}
\textbf{Conclusion.} In this paper, we experimentally demonstrate that SNNs trained with direct encoding and BPTT can undergo catastrophic model collapse when hetero-training which is common in real‑world scenarios. Through theoretical analysis, we show that the repeated inputs of direct encoding combined with gradient accumulation in BPTT induce extremely large spectral radius in the Hessian matrix of the loss function, causing the model parameters to become trapped in precarious local minima. Motivated by these insights, we propose a hyperparameter-free method named Domain Eigencomponent Projection (DEP): by orthogonally projecting gradients to precisely eliminate their dominant components, DEP effectively reduces the Hessian spectral radius. Extensive evaluations under both homogeneous and heterogeneous training conditions demonstrate that DEP substantially enhances SNN robustness, paving the way for safer and more reliable deployments.
 
\textbf{Limitation.} The gradient‐based adjustment inherent to DEP induces a deliberate divergence between the gradients actually applied during training and those that would be obtained under an ideal, unmodified regime. While this adjustment markedly bolsters SNNs' robustness against adversarial attacks, it unavoidably incurs a slight degradation in accuracy when evaluated on unperturbed data. This limitation is also prevalent in existing SOTA methods \cite{ding2024enhancing, ding2024robust, geng2023hosnn, xu2024feel} according to Table \ref{tab1}.

\textbf{Broader Impact.} (i). Firstly, DEP is hyperparameter‑free, which streamlines deployment by eliminating the need for costly tuning and ensures seamless scalability. (ii). Secondly, DEP introduces only a negligible computational overhead during training, consisting of a single SVD and gradient modification per batch. Moreover, DEP incurs no extra computational cost during inference. (iii). Furthermore, DEP mitigates vulnerability in hetero-training has profound practical importance. Imagine an attacker with access to only a single batch of samples: if those adversarial examples were scattered randomly across the dataset, their induced distributional shift would likely be absorbed and flattened, limiting the attack’s impact. However, by concentrating the adversarial examples within one batch, the adversary can exploit an inherent SNN vulnerability to trigger catastrophic model collapse. Our experiments demonstrate that DEP can substantially mitigate the occurrence of such model collapse. In summary, DEP delivers an effective, highly scalable defense method that enables safer and more reliable deployment of SNNs.



\bibliographystyle{plain} 
\bibliography{reference}

\newpage
\appendix

\section{Adversarial Attack Details and Configurations}
\label{App_attack}

\textbf{FGSM.} FGSM is a simple yet effective technique to generate adversarial examples. In FGSM, given an input \( x \) with its true label \( y \), a perturbation is computed in the direction of the gradient of the loss function with respect to \( x \). The perturbation is defined as:

\begin{equation}
    \delta = \epsilon \cdot \text{sign}\left( \nabla_x \mathcal{L}(f(x), y) \right).
\end{equation}

where \( \epsilon \) controls the magnitude of the perturbation and \( h(x) \) represents the model's output. The adversarial example is then constructed as \( x + \delta \), which is designed to force the model into misclassification.

\textbf{PGD.}  PGD is an iterative method for generating adversarial examples and can be regarded as an extension of the FGSM. PGD updates the adversarial example iteratively by performing a gradient ascent step and then projecting the result back onto the feasible set defined by the \( L_p \)-norm constraint. Formally, the update rule is given by:
\begin{equation}
    x^{t+1} = \Pi_{\mathcal{B}(x, \epsilon)}\Bigl(x^t + \alpha \cdot \operatorname{sign}\bigl(\nabla_x \mathcal{L}(f(x^t), y)\bigr)\Bigr).
\end{equation}

where \( \alpha \) denotes the step size, \(\mathcal{L}(f(x^t), y)\) is the loss function of the model \( h \) with true label \( y \), and \( \Pi_{\mathcal{B}(x, \epsilon)} \) is the projection operator that projects the perturbed example back into the ball \(\mathcal{B}(x, \epsilon) = \{ x' : \|x' - x\|_p \le \epsilon \}\). By iterating this process, PGD effectively seeks a perturbation that maximizes the loss while ensuring that the adversarial example remains within the specified perturbation budget.

For these attack methods, we set $\epsilon=8/255$ for all experimental cases. For PGD, step size $\alpha=0.01$ and step number $K=7$.

\section{Experimental Implementation}
\label{App_exp}

In our experiments, all training cases are implemented using PyTorch \cite{paszke2019pytorch} with the SpikingJelly \cite{fang2023spikingjelly} framework and executed on an NVIDIA GeForce RTX 5090 GPU. For each dataset, we utilize the hyperparameters listed as Table 5, consistently employing the SGD optimizer and setting the number of time steps to 4 and the membrane time constant $\tau$ to 1.1. We leverage the PyHessian framework \cite{yao2020pyhessian} to compute Hessian eigenvalues.

\begin{table*}[h]
    \centering
    \label{tab_hyper}
    \renewcommand{\arraystretch}{1.2} 
    \caption{Hyperparameter settings for experiments.}
    \resizebox{1\textwidth}{!}{
    \begin{tabular}{>{\centering\arraybackslash}m{2.5cm}>{\centering\arraybackslash}m{3cm}>{\centering\arraybackslash}m{2cm}>{\centering\arraybackslash}m{2cm}>{\centering\arraybackslash}m{2cm}>{\centering\arraybackslash}m{2cm}}
    
    \noalign{\hrule height 1.5pt}

    \cellcolor{gray!20}Dataset &  \cellcolor{gray!20} Model  & \cellcolor{gray!20} LeaningRate &  \cellcolor{gray!20} WeightDecay &  \cellcolor{gray!20} Epoch &  \cellcolor{gray!20} BatchSize \\
    
    \hline
    
    CIFAR-10 & VGG-11 & 0.1 & 5e-5 & 300 & 128 \\
    CIFAR-100 & VGG-11 & 0.1 & 5e-4 & 300 & 128 \\
    TinyImageNet & VGG-16 & 0.1 & 5e-4 & 300 & 128 \\
    ImageNet & NF-ResNet-18 \cite{brock2021high} & 0.1 & 1e-5 & 100 & 512 \\
    ImageNet (AT) & ResNet-18  & 0.1 & 1e-5 & 100 & 512 \\

    \noalign{\hrule height 1.5pt}
    \end{tabular}
    }
\end{table*}

\section{Detailed Experimental Result}
\label{App_exp_res}

In this section, we present the full experimental data underlying Figures 3 and 4 from the main text.

\begin{table}[]
    \centering
    \label{App_tab_exp1}
    \renewcommand{\arraystretch}{1.2} 
    \caption{Performance of DEP with different attack methods (\%).} 
    
    \resizebox{1\textwidth}{!}{
    \begin{tabular}{>{\arraybackslash}m{2.5cm}|>{\centering\arraybackslash}m{0.05cm}|>{\centering\arraybackslash}m{1.2cm}>{\centering\arraybackslash}m{1.2cm}>{\centering\arraybackslash}m{1.2cm}>{\centering\arraybackslash}m{1.2cm}>{\centering\arraybackslash}m{1.2cm}>{\centering\arraybackslash}m{1.2cm}>{\centering\arraybackslash}m{1.2cm}>{\centering\arraybackslash}m{1.2cm}>{\centering\arraybackslash}m{1.2cm}}
    
    \noalign{\hrule height 1.5pt}

    \centering\cellcolor{gray!20}\textbf{Attack} & & \cellcolor{gray!20}$\epsilon=0$ & \cellcolor{gray!20}2 &\cellcolor{gray!20} 4 &\cellcolor{gray!20} 6 &\cellcolor{gray!20} 8 &\cellcolor{gray!20} 16 &\cellcolor{gray!20} 32 &\cellcolor{gray!20} 64 & \cellcolor{gray!20} 128 \\
    
    \hline
    \multicolumn{11}{c}{\cellcolor{app1}\textbf{CIFAR-10}}\\
    \hline
    
    SNN FGSM WB& & 93.75 & 17.06 & 14.22 & 11.78 &  8.19 & 4.21 & 1.99 & 0.58 & 0.00 \\
    SNN FGSM BB& & 93.75 & 24.13 & 18.41 & 13.86 & 10.26 & 7.88 & 3.82 & 1.48 & 0.59 \\
    SNN PGD WB& & 93.75 &  2.37 &  1.01 &  0.34 &  0.03 & 0.00 & 0.00 & 0.00 & 0.00 \\
    SNN PGD BB& & 93.75 &  4.01 &  2.80 &  1.20 &  0.89 & 0.02 & 0.00 & 0.00 & 0.00 \\
    \cline{1-1} \cline{3-11}
    
    DEP FGSM WB & & 90.21   & 59.75 & 58.22 & 57.00 & 55.86& 50.36 & 39.82 & 26.61 & 8.01 \\
    DEP FGSM BB & &  90.21  & 81.42& 79.37 & 76.65 & 74.90 & 67.18 & 49.30 & 31.09 & 12.81 \\
    DEP PGD WB & &  90.21  & 40.78 & 37.46 & 34.50 & 31.44 & 22.04 & 6.36 & 0.03 & 0.00 \\
    DEP PGD BB & &  90.21  & 57.88 & 54.09 & 47.80 & 43.15 & 30.73 & 12.11 & 0.91 & 0.24 \\

    \hline
    \multicolumn{11}{c}{\cellcolor{app2}\textbf{CIFAR-100}}\\
    \hline

    SNN FGSM WB& & 72.39 &  9.37 &  7.42 &  5.46 &  4.55 &  2.35 &  1.19 &  0.33 & 0.00 \\
    SNN FGSM BB& & 72.39 & 13.26 & 12.11 & 10.84 &  9.16 &  5.31 &  2.07 &  1.47 & 0.49 \\
    SNN PGD WB& & 72.39 &  2.53 &  1.15 &  0.50 &  0.19 &  0.02 &  0.00 &  0.00 & 0.00 \\
    SNN PGD BB& & 72.39 &  3.65 &  2.68 &  1.89 &  0.78 &  0.15 &  0.02 &  0.00 & 0.00 \\
    \cline{1-1} \cline{3-11}
    
    DEP FGSM WB & & 70.26 & 27.89 & 26.65 & 24.83 & 23.81 & 14.81 & 8.38 & 3.36 & 0.20 \\
    DEP FGSM BB & & 70.26 & 40.89 & 38.67 & 35.67 & 31.55 & 21.63 & 13.31 & 9.03 & 1.01 \\
    DEP PGD WB & & 70.26 & 20.52 & 15.78 & 12.35 & 8.09  & 2.37  & 0.00 & 0.00 & 0.00 \\
    DEP PGD BB & & 70.26 & 27.99 & 26.01 & 23.84 & 18.37 & 9.01  & 1.07  & 0.00 & 0.00 \\

    \hline
    \multicolumn{11}{c}{\cellcolor{app3}\textbf{TinyImageNet}}\\
    \hline

    SNN FGSM WB& & 56.82 &  9.42 &  7.60 &  4.82 &  3.51 &  1.53 &  0.67 &  0.00 &  0.00 \\
    SNN FGSM BB& & 56.82 & 16.27 & 15.59 & 13.69 & 12.46 &  8.20 &  4.80 &  2.74 &  1.28 \\
    SNN PGD WB& & 56.82 &  2.98 &  1.46 &  0.89 &  0.14 &  0.00 &  0.00 &  0.00 &  0.00 \\
    SNN PGD BB& & 56.82 &  4.07 &  2.99 &  2.12 &  1.67 &  0.99 &  0.46 &  0.00 &  0.00 \\
    \cline{1-1} \cline{3-11}
    
    DEP FGSM WB & & 54.54 & 22.80 & 21.74 & 21.00 & 19.50 & 12.66 & 6.58  & 1.58  & 0.02 \\
    DEP FGSM BB & & 54.54 & 34.63 & 31.02 & 29.40 & 27.68 & 20.22 & 10.35 & 3.20  & 1.38 \\
    DEP PGD WB & & 54.54 & 18.42 & 15.70 & 13.22 & 12.02 & 5.78  & 2.96  & 0.20  & 0.00 \\
    DEP PGD BB & & 54.54 & 24.92 & 20.66 & 16.82 & 14.84 & 6.55  & 4.47  & 2.10  & 0.25 \\

    \hline
    \multicolumn{11}{c}{\cellcolor{app4}\textbf{ImageNet}}\\
    \hline

    SNN FGSM WB& & 57.84 & 10.75 &  8.59 &  6.73 &  4.99 &  1.56 &  0.14 &  0.01 & 0.00 \\
    SNN FGSM BB& & 57.84 & 12.13 & 11.43 & 10.25 &  8.46 &  5.42 &  3.93 &  1.25 & 0.35 \\
    SNN PGD WB& & 57.84 &  1.02 &  0.48 &  0.13 &  0.01 &  0.00 &  0.00 &  0.00 & 0.00 \\
    SNN PGD BB& & 57.84 &  9.19 &  6.05 &  4.06 &  3.67 &  2.02 &  1.79 &  0.05 & 0.00 \\
    \cline{1-1} \cline{3-11}
    
    DEP FGSM WB & & 53.47 & 18.62 & 17.31 & 15.65 & 14.69 & 10.98 & 5.30  & 2.01  & 0.02 \\
    DEP FGSM BB & & 53.47 & 27.68 & 25.58 & 24.44 & 22.40 & 16.24 & 8.96  & 4.90  & 2.16 \\
    DEP PGD WB & & 53.47 & 11.63 & 8.12  & 6.33  & 4.30  & 2.97  & 1.50  & 0.00  & 0.00 \\
    DEP PGD BB & & 53.47 & 16.10 & 12.22 & 8.33  & 5.68  & 4.09  & 1.67  & 0.20  & 0.04 \\
    
    \noalign{\hrule height 1.5pt}
    \end{tabular}
}
\end{table}

\begin{table}[]
    \centering
    \label{App_tab_exp2}
    \renewcommand{\arraystretch}{1.2} 
    \caption{Performance comparison with different PGD step number on CIFAR-10 (\%).} 
    
    \resizebox{1\textwidth}{!}{
    \begin{tabular}{>{\centering\arraybackslash}m{2.5cm}|>{\centering\arraybackslash}m{0.05cm}|>{\centering\arraybackslash}m{1.1cm}>{\centering\arraybackslash}m{1.1cm}>{\centering\arraybackslash}m{1.1cm}>{\centering\arraybackslash}m{1.1cm}>{\centering\arraybackslash}m{1.1cm}>{\centering\arraybackslash}m{1.1cm}>{\centering\arraybackslash}m{1.1cm}>{\centering\arraybackslash}m{1.1cm}>{\centering\arraybackslash}m{1.1cm}>{\centering\arraybackslash}m{1.1cm}}
    
    \noalign{\hrule height 1.5pt}

    \cellcolor{gray!20}\textbf{Method} & & \cellcolor{gray!20}$K=7$ & \cellcolor{gray!20}10 &\cellcolor{gray!20} 15 &\cellcolor{gray!20} 20 &\cellcolor{gray!20} 30 &\cellcolor{gray!20} 40 &\cellcolor{gray!20} 50 &\cellcolor{gray!20} 60 & \cellcolor{gray!20} 70 & \cellcolor{gray!20} 80\\
    
    \cline{1-1} \cline{3-12}
    
    SNN & & 0.03 & 0.01 & 0.00 & 0.00 & 0.00 & 0.00 & 0.00 & 0.00 & 0.00 & 0.00 \\
    SNN (AT) & & 14.07 & 13.48 & 12.99 & 12.03 & 11.36 & 10.79 & 10.49 & 10.44 & 10.42 & 10.41 \\
    DEP & & 31.44 & 30.34 & 29.12 & 27.66 & 26.40 & 25.74 & 25.50 & 25.46 & 25.45 & 25.44 \\
    DEP (AT) & & 44.38 & 43.10 & 42.56 & 41.29 & 40.78 & 40.01 & 39.80 & 39.78 & 39.77 & 39.77 \\

 \noalign{\hrule height 1.5pt}
    \end{tabular}
}
\end{table}

\end{document}